# BTPD: A Multilingual Hand-curated Dataset of Bengali Transnational Political Discourse Across Online Communities

DIPTO DAS, University of Toronto, Canada

SYED ISHTIAQUE AHMED, University of Toronto, Canada

SHION GUHA, University of Toronto, Canada

Understanding political discourse in online spaces is crucial for analyzing public opinion and ideological polarization. While social computing and computational linguistics have explored such discussions in English, such research efforts are significantly limited in major yet under-resourced languages like Bengali due to the unavailability of datasets. In this paper, we present a multilingual dataset of Bengali transnational political discourse (BTPD) collected from three online platforms, each representing distinct community structures and interaction dynamics. Besides describing how we hand-curated the dataset through community-informed keyword-based retrieval, this paper also provides a general overview of its topics and multilingual content.

## 1 Introduction

Computer-mediated communication in online communities profoundly shapes contemporary political discourse. Prior computer-supported cooperative work (CSCW) research has studied the discussions of political issues with one's peers in the context of Bengali communities, especially focusing on their postcolonial conditions, decolonial efforts, and intriguing cultural and political dynamics in the Global South [15, 18, 21]. Often dubbed as "adda," such political discourse is a "something quintessentially Bengali, ... an indispensable part"of Bengali practices [9]. The Bengali people are the third largest ethnic group in the world [4], native to South Asia. Through the postcolonial partition of the region, Bengali communities were divided between Bangladesh and India [11], particularly in the states of West Bengal, Assam, and Tripura. Such a colonial formation of transnational dynamics among the Bengali communities makes the political perceptions and discourse historically complex [10]. The region has also become geopolitically important in recent times due to its strategic position in South Asian trade and connectivity, migration dynamics, and being situated within and near emerging global powers (e.g., China, India) and major stakeholders in many global issues (e.g., climate change) [2, 27]. Moreover, with around half a million Bangladeshi Bengalis living in the US[1, 63], as well as many Indian Bengalis who often identify as Indian-Americans, and with growing Bengali communities in major Canadian cities [8], the influence of such large ethnic enclaves on North American politics and the economy is steadily growing [45].

CSCW community often develops datasets of computer-mediated political discussions and empirically studies those interactions [32, 62, 66]. As contemporary political discourse among the Bengalis frequently takes place online [15], a dataset of their political discourse would enable CSCW researchers to study its transnational dynamics, complexities, and significance. However, despite Bengali being the sixth-largest native language [42] and having a strong web presence, there exist fewer resources in different linguistic data sites and consortia for Bengali than for other major languages [40]. Given their diverse backgrounds, Bengali communities adopt different online platforms based on platform-specific affordances, i.e., perceived and actual interaction possibilities, and tailor their political discussions accordingly. In this paper, we develop a dataset[1] of Bengali transnational political discourse (BTPD) with multilingual support, collected from three online platforms of different community structures and objectives, namely Reddit, Politics Stack Exchange[2], and Bengali Quora[3]. In the following sections, we will explain ways to conceptualize different types of online communities and review existing research, outline our data collection, preprocessing, and organization strategies, and describe the dataset using natural language processing (NLP) methods.

---

[1] Publicly available upon the paper's acceptance.    [2] https://politics.stackexchange.com/    [3] https://bn.quora.com/

## 2  Literature Review

While most existing research on political discussions online focuses on online platforms, such as Twitter and Facebook [14, 34, 39] that are typically understood as "social media," Bruckman argues that these online communities should be viewed as a prototype-based category [6], defined not by rigid inclusion and exclusion rules, but by their prototypical members. Though social media platforms are more representative of online communities, which are often riddled with political misinformation [65], Question-and-Answer (Q&A) platforms, which usually offer better information quality with adequate support for fostering connections among users [26], can also be viewed as online communities. Moreover, the degree to which a platform embodies the prototypicality of a community can be viewed as a cultural construct [6]. For example, while prior studies on Quora focused on collective wisdom, reputation, quality of answers−objectives that are typical for Q&A sites [52, 53, 67], Das and colleagues demonstrated how the Bengali users from Bangladesh and India fostered transnational communities based on their linguistic and cultural similarities and participated in sociopolitical discussions through this platform [19, 21]. While most political discourse datasets rely on news sources or single platforms [49], that may not reflect the holistic online political discourse. Considering that different platforms facilitate user interaction differently, any datasets on political discourse should be curated across multiple sites.

Similar to many other fields adjacent to CSCW, such as human-computer interaction (HCI), NLP, and algorithmic fairness [40, 44, 58], most research and resources in computational social science, for example, in studying political discourse, overwhelmingly focus on the Global North contexts [7, 12, 23, 33, 38, 43]. With recent studies focused on discourse in the Global South, countries like India, Brazil, Indonesia, and Nigeria [25, 47, 48, 50] have facilitated cross-cultural analyses of global political participation [41, 51]. However, there exists a dearth of datasets for studying the transnational Bengali ethnolinguistic communities in Bangladesh and India. While most NLP datasets in this under-resourced language have focused on tasks like sentiment analysis and hate speech detection [22, 54, 56], some recent datasets have focused on bias evaluation [16, 17], Q&A [60], machine translation [31], etc. Though some recent single platform-sourced datasets of public opinion in Bengali exist [13, 30], they primarily feature product reviews or discussions on global events rather than political discourse directly relevant to the Bengali people. Moreover, most Bengali datasets are shaped by the construct of nationality, framing it either as a language specific to Bangladesh or as a regional language in India. This paper seeks to address this gap by foregrounding the transnational Bengali political discourse in Bangladesh and India. Following recommendations in human-centered data science [3, 36] and common practices in HCI and CSCW [20, 61], we hand-curated the dataset from multiple online communities through keyword search.

## 3  Dataset Creation

### 3.1  Choice of Platforms

Drawing on Bruckman's argument that community is best understood as a prototype-based category [6], we chose three online platforms that exhibit varying degrees of different prototypicality as communities. For our paper on preparing a corpus of Bengali political discussions online, we collected political discussions from Reddit, the Politics Forum on Stack Exchange (PoliticsSE), and Bengali Quora (BnQuora). Among these, Reddit aligns most closely with traditional notions of community due to its persistent user identities, subreddit-based governance structures, and ongoing interactions centered around shared interests [68]. In contrast, though users build reputations and some expert-driven communities develop around specific topics on Stack Exchange, this platform prioritizes high-quality information exchange through structured Q&A rather than sustained interaction for social bonding [55] and, thus, is a less prototypical example of a community. Compared to these two, BnQuora falls somewhere in between. Though it has a looser sense of community compared to Reddit, as discussions in Q&A threads are more individual-driven than group-based, [19] have found its effectiveness in fostering a sense of social relationship among Bengali users from different regions based on their cultural similarities, reinforcing Bengali ethnolinguistic identity, and facilitating political discourses.

## 3.2 Data Collection

We collected data from these platforms through keyword searches. The list of keywords and data collection process varied across platforms based on their technical scaffolds and topical focus.

*3.2.1  Reddit.* Reddit facilitates decentralized discussions through subreddits, which are often geographically anchored or based on similar cultures and interests. Therefore, we could look for subreddits related to politics and Bengali contexts. Since most politics-related subreddits (e.g., r/politics) are US-centered or strongly guided by US-adjacency (e.g., r/Ask_Politics) as found by [28], to collect posts on Bengali politics, following [21], we included the subreddits related to the geographic regions where Bengali people live (r/bangladesh, r/westbengal) and their political centers (r/Dhaka, r/kolkata) and their shared linguistic backgrounds (r/bengalilanguage) as communities where discussions on politics in Bengali social contexts are likely to occur,. These subreddits are moderated and use flairs (e.g., "Seeking advice/পরামর্শ") that indicate the type and topic of the content. For our data collection, we looked for posts in those subreddits that used the flairs: "Politics/রাজনীতি", "Discussion/আলোচনা," and "News/সংবাদ."

We used the Python Reddit API Wrapper (PRAW) to collect data from January 25, 2025, to February 17, 2025. This period reflects the code execution time, not the posting times of the posts. We collected the posts' titles, URLs, bodies, flair, times of posting, and comments. While Reddit employs a nested branching structure for comments, we stored the comments as a flat list. Based on our long membership in the previously mentioned subreddits, we have observed certain differences in how flairs are used in various subreddits. After data collection, we similarly noticed how different subreddits used flair more or less frequently to indicate political posts and how the same flair in various subreddits resulted in differing numbers of political posts as well as posts unrelated to politics. For example, political posts in r/bangladesh often bear the flair "Politics/রাজনীতি", whereas r/westbengal uses the flair "News/সংবাদ" and uses the flair "Politics/রাজনীতি" less frequently. In both subreddits, the flair "Discussion/আলোচনা" is used in posts related to politics as well as other topics. Hence, to keep the corpus relevant to Bengali political discussions, we excluded posts on other topics (e.g., posts bearing the "Discussion/আলোচনা" flair but focusing on different topics). Table 1 lists the subreddits and numbers of members and collected posts.

Table 1. Subreddits and their number of members (top x% of largest communities on Reddit) and political posts from there included in our dataset.

| Subreddit | #Members | #Political posts |
| --- | --- | --- |
| r/bangladesh | 75K (2%) | 601 |
| r/westbengal | 5.5K (10%) | 206 |
| r/Dhaka | 54K (3%) | 309 |
| r/kolkata | 331K (1%) | 49 |
| r/bengalilanguage | 26K (4%) | 55 |

*3.2.2  PoliticsSE.* In contrast to the diverse topics discussed in our selected subreddits, PoliticsSE is a Q&A forum solely for political discourse, ensuring the inherent topical relevance of its posts. As such, for our data collection, we can prioritize the contextual relevance of the data to Bengali communities rather than concerns about the broader topical focus. We retrieved PoliticsSE's latest data dump from the Internet Archive, which includes data from the platform's launch until December 31, 2024. As before, we used the keywords mentioning the regions where the Bengali people are native (e.g., `Bangladesh`, `West Bengal`), their political centers (e.g., `Dhaka`, `Kolkata`) and their language and community name (`Bengali`) to identify posts on PoliticsSE related to the context of Bengali communities based on their titles, bodies, and tags. We manually read through the posts and only retained the unique posts while excluding the ones not directly related to political discussion in the Bengali context (e.g., posts that mention Bangladesh as a passing example while generally discussing different parliamentary structures around the world). We also retained their metadata, such as URLs and posting time. Table 2 shows the number of posts identified using keywords from the PoliticsSE data dump and the ones relevant to Bengali politics.

Table 2. Keywords, number of posts mentioning those keywords, and number of posts relevant to Bengali political discussion among those identified posts.

| Keyword | #Posts identified through keywords | #Posts identified as relevant |
|---|---|---|
| Bangladesh | 234 | 209 |
| West Bengal | 14 | 14 |
| Dhaka | 10 | 10 |
| Kolkata | 5 | 5 |
| Bengali | 19 | 17 |

*3.2.3 BnQuora.* Quora's Q&A structure, which encourages diverse perspectives on a given topic, fosters in-depth discussions on controversial subjects, including politics [35, 64]. Similarly, as described in the previous section, BnQuora provides a unique space for in-depth analysis of Bengali political discourse without significant concerns about the contextual or broader topical relevance. Hence, instead of searching with keywords on the broad topic (e.g., politics), Das et al. [20] recommended using more specific terms related to the broader topic to identify Q&A threads to collect data from BnQuora. We conducted a Qualtrics survey to know what specific topics are crucial to contemporary Bengali political discourse. The survey presented common political discussion points [46] as options while allowing participants to add unlisted responses. We circulated the survey through our social networks as members of Bengali communities in Bangladesh and West Bengal and through Bangladeshi, Indian, and South Asian student organizations at two North American universities. We thematically consolidated the 74 responses received between October 5 and 21, 2024, into a list of key topics/themes in Bengali political discussions and used the ten most prominent ones to collect our dataset. Besides these topics, we used other related keywords (see Table 3) to search for Q&A threads on BnQuora. We periodically ran a Python script using Selenium from November 1, 2024, to February 15, 2025, to automate browser interactions (e.g., refresh, scroll) to manage dynamic page content, which collected Q&A threads containing those keywords.

## 3.3 Data Preprocessing and Organization

Our collected data from PoliticsSE and BnQuora are primarily in English and Bengali, respectively, with occasional use of the other language for certain terms or phrases. However, the languages of Reddit data vary significantly, including Bengali and English, with occasional code-switching and Romanized Bengali, i.e., phonetic Bengali using English fonts. We translated all collected posts in Bengali and English using OpenAI's API with the GPT-4 engine, which is comparable to commercial translation products [37], using the following prompts: *"You are a translator who can translate* {`Bengali/English`} *and Banglish (Bengali in romanized fonts) to* {`English/Bengali`}*."* In our capacity as natively Bengali-speaking researchers, we also manually verified and fixed the translations if needed. Hence, for each unique post ID, besides the original post, which may have used a mix of languages, we have its translations in Bengali and English. This makes our dataset uniform and multilingual, including a total of 2235 posts' original titles and bodies, their translations in Bengali and English, answers and comments, posting time, and tags, if available. Though the concern of misinformation is often intensified in political discussions and our dataset comes from online communities with varied information quality, where StackExchange platforms like PoliticsSE are seen as reliable [55] but Reddit has documented misinformation issues [5, 59], not screening for misinformation while including a post in our dataset allowed BTPD to stay true to the dynamics and reflect the nature of political discussions in Bengali communities online.

Table 3. Key topics of Bengali political discussions, related keywords, and number of Q&A threads collected.

| Topical Themes | Keywords | #Q&A threads |
|---|---|---|
| foreign policy | পররাষ্ট্র নীতি | 140 |
| constitution | সংবিধান | 246 |
| secularism | ধর্মনিরপেক্ষতা | 57 |
| public education | সরকারি শিক্ষাক্রম | 18 |
| cultural identity | বাঙালি, বাংলাদেশি, বাংলাদেশী, ভারতীয়, ইন্ডিয়ান | 87 |
| LGBTQ+ rights | সমকামী/ট্রান্সজেন্ডার অধিকার | 42 |
| political parties | আওয়ামী লীগ, জামায়াতে ইসলাম, তৃণমূল কংগ্রেস, বিএনপি, বিজেপি | 119 |
| religion | ধর্ম, মুসলিম, ইসলাম, হিন্দু, ধর্মীয় সংখ্যালঘু | 29 |
| women's rights | নারী অধিকার | 131 |
| ethnic minorities | আদিবাসী, ক্ষুদ্র নৃতাত্ত্বিক জনগোষ্ঠী | 7 |

## 4 Dataset Content

In this section, we provide a brief descriptive overview of our developed dataset. After pre-processing (e.g., excluding stopwords, stemming), Table 4 shows that lengths (average and median) and timestamps of the earliest and the latest posts in our dataset varied significantly across platforms.

Table 4. Overview of the collected data by platforms.

| Platform | #Sentences | #Words | Earliest and latest posts |
|---|---|---|---|
| Reddit | 37.1, 15.0 | 198.0, 99.5 | 2023/01/02, 2025/02/17 |
| PoliticsSE | 16.2, 10.0 | 130.3, 90.0 | 2012/12/13, 2024/12/22 |
| BnQuora | 23.7, 12.0 | 213.4, 109.0 | 5 years ago, 2025/02/15 |

As described earlier, our multilingual dataset includes the Bengali and English versions of each post. To compare the variances in the Bengali and English versions of the posts, we used principal component analysis (PCA) on their TF-IDF (Term Frequency-Inverse Dense Frequency) vectors. Examining the differing elbows in the scree plot (see Figure 1(a)), we can see that the number of principal components needed to retain a fixed proportion of variance (e.g., 80%) varies across languages–for instance, approximately the first 500 for Bengali and 1000 for English.

Figure 1(b) shows the common words appearing in our dataset using a wordcloud. Though we addressed language-specific characteristics (e.g., Bengali's bidirectional structure) and provided Unicode fonts, the existing NLP tools could not visualize the Bengali wordcloud properly. We conducted topic modeling of the titles and bodies of the posts to get an overview of the broad topics included in our dataset. Given the lack of enough evidence of how common topic modeling approaches like latent Dirichlet allocation (LDA) and non-negative matrix factorization (NMF) work in Bengali, we tried to identify topics through clustering of the posts based on their sentence embeddings but did not find this approach informative. Since NMF works better than other common approaches like LDA for topic modeling of short texts [24, 29], in Table 5, we report ten topics identified by NMF on the posts' English translations with each topic's corresponding top five words.

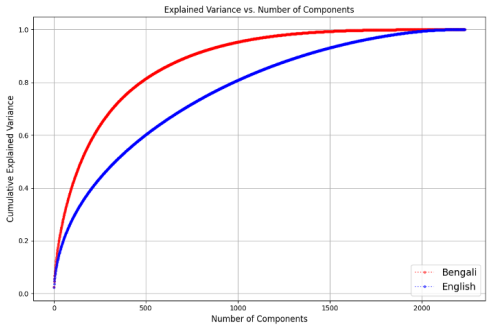 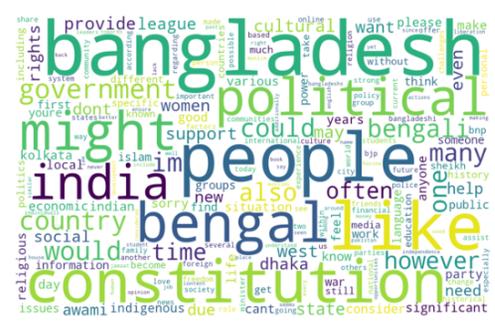

Fig. 1. (a) Scree plots of principal component analysis of Bengali (in red) and English (in blue) versions of the posts in our dataset (b) Common words in the English translations of the posts

Table 5. Topics identified in the English versions of the posts by NMF with common words.

| Topic | Words | Topic | Words |
|---|---|---|---|
| 0 | assist, sorry, request, information, content | 1 | country, like, people, Awami-League, time |
| 2 | constitution, according, written, country, Indian | 3 | West-Bengal, chief-minister, BJP, Mamata-Banerjee, state |
| 4 | India, foreign-policy, Dr-Ambedkar, Hindu, draft | 5 | Indigenous, people, communities, tribes, Bengalis |
| 6 | provide, text, translation, information, need | 7 | women-rights, men, Islam, equal, freedom |
| 8 | Bengali, Trinamool, Congress, BJP, parties | 9 | Bangladesh, secularism, Pakistan, war, prime-minister |

While some of these topics (e.g., 0, 6) are generic, some topics closely relate to particular domains of Bengali political discourse. For example, topics 3 and 8 focus on West Bengal's state-level politics in India, whereas topic 9 covers Bangladesh's historical political issues, and topic 5 deals with the politics around settler Bengalis and the Indigenous and ethnic tribes in Bangladesh. Interestingly, topic 7 seems to engage closely with equality of rights and freedom across different genders in Islam. Overall, topics 4, 7, and 9 highlight the centrality of religion and caste to politics in Bengal by mentioning words like Islam, secularism, Hindu, and Dr. Ambedkar (an Indian social reformer with great contributions in alleviating underprivileged castes, who, being elected from the Bengal region, chaired the Indian constitution drafting committee [57]). While some of the top words shown in Table 5 are identical to words used for keyword-based search, NMF surfaced more important keywords and identified connections among the words that were not implied during data collection.

## 5 Conclusion

This poster follows traditional NLP strategies while being informed by CSCW and social computing scholarship in considering different prototypical examples of online communities. It develops a textual corpus of transnational Bengali political discussions, which would address a resource need in one of the major global languages and be useful for examining cross-platform information dynamics and cultural and longitudinal shifts in political discourse in one of the largest global ethnolinguistic communities. While future research should contribute more data instances to BTPD from other online communities, include additional metadata like fact-checking labels, and link the online discussions with reliable sources, this artifact would facilitate political deliberation among Bengali communities and critical algorithmic audits of political biases of Bengali NLP systems, such as large language models, automated content moderation, and recommendation systems.